\documentclass[10pt,twocolumn,letterpaper]{article}

\usepackage{wacv}
\usepackage{times}
\usepackage{epsfig}
\usepackage{graphicx}
\usepackage{amsmath}
\usepackage{amssymb}
\usepackage{microtype}
\usepackage{subfigure}
\usepackage{amsfonts}       
\usepackage{nicefrac}       
\usepackage{amsthm}
\DeclareMathOperator*{\argmax}{argmax}
\usepackage[retainorgcmds]{IEEEtrantools}
\usepackage{comment}
\graphicspath{{latex/images/}}

\wacvfinalcopy 

\def\wacvPaperID{341} 

\ifwacvfinal\fi
\begin{document}

\title{An Universal Image Attractiveness Ranking Framework}

\author{Ning Ma \hspace{0.2 cm} Alexey Volkov\thanks{Currently a software engineer at Google} \hspace{0.2cm} Aleksandr Livshits \hspace{0.2cm} Pawel Pietrusinski \hspace{0.2cm} Houdong Hu \hspace{0.2cm} Mark Bolin\\
Microsoft\\
{\tt\small \{ninm, alevol, alliv, Pawel.Pietrusinski, houhu, markbo\}@microsoft.com}
}

\maketitle
\ifwacvfinal\thispagestyle{empty}\fi
\begin{abstract}
We propose a new framework to rank image attractiveness using a novel pairwise deep network trained with a large set of side-by-side multi-labeled image pairs from a web image index. The judges only provide relative ranking between two images without the need to directly assign an absolute score, or rate any predefined image attribute, thus making the rating more intuitive and accurate. We investigate a deep attractiveness rank net (DARN), a combination of deep convolutional neural network and rank net, to directly learn an attractiveness score mean and variance for each image and the underlying criteria the judges use to label each pair. The extension of this model (DARN-V2) is able to adapt to individual judge's personal preference. We also show the attractiveness of search results are significantly improved by using this attractiveness information in a real commercial search engine. We evaluate our model against other state-of-the-art models on our side-by-side web test data and another public aesthetic data set. With much less judgments (1M vs 50M), our model outperforms on side-by-side labeled data, and is comparable on data labeled by absolute score.
\end{abstract}

\section{Introduction}
While object classification and detection have achieved performance comparable to human, the image attractiveness evaluation is a far from solved problem. Image attractiveness plays a critical role in many multimedia applications such as image retrieval in Bing/Google/Pinterest, etc. For example, when a customer issues a query, a search engine should deliver images as attractive as possible while not impairing relevance.

Substantial progress has been made in evaluating image aesthetic value, including image feature extraction, predictive models  and data set collections. Early feature extraction focused on hand-crafted features by utilizing predefined aesthetic attributes, low level image statistics, generic content features or high-level cue based rules ~\cite{Datta06, Dhar11,  Y06, Luo11, Luo08, Marchesotti08,  Nishiyama11}. Recently, benefiting from the fast advance of deep learning in object recognition, deep convolutional neural networks (CNN) have been widely used in image attractiveness prediction and obtained impressive performance \cite{Bianco2017, Bosse2016, Jin2016, Kang14, Kao2015, Kong2016, Lu2014, Lu2015, Ma2017,Mai2016, Talebi2017,Xue13}. At the same time, more image attractiveness related data sets have been released, thus making this deep neural network approach more feasible \cite{Kong2016,Larson2010, Murray2012,Sheikh2005, Talebi2017}.

However, most previous work formalizes attractiveness evaluation as a regression or classification problem to predict human rated attractiveness score. This formalization arises naturally due to the fact that existing data sets usually require judges to give an absolute score to each image, e.g. from 1 to 10 in the Aesthetic Visual Analysis (AVA) data set \cite{Murray2012}. Nevertheless, unlike classification problem, judging image attractiveness is subjective and very challenging. Giving an absolute score to each image is equal to requiring each judge to directly rank hundreds of thousands of images, which is a very difficult task and usually requires hundreds of sophisticated judges as in the AVA data. Some data sets, like Aesthetics And Attributes
Database (AADB)~\cite{Kong2016}, require labeling on predefined aesthetic attributes such as VividColor, ShallowDOF, etc. These kind of judgments become even more challenging in image retrieval due to the extremely diverse and large number of images in the search index. A recent work \cite{Kong2016} takes this issue into account by incorporating a rank related loss into the the regression loss when training their model using absolute score rated images. To a human judge, answering which image looks more appealing when compared side by side is a much easier and less subjective task compared to directly assigning an absolute score to an image. Ideally, we would build a sophisticated machine learning model to learn an attractiveness representation for each image by utilizing the pairwise-ranked image pairs. In this way, we leave a relatively simple task to the human judges and let the computer solve the harder problem.

In this paper, we propose an alternative framework to rank image attractiveness using a novel model trained with a large set of side-by-side multi-labeled image pairs. We collect a large and diverse set of image pairs from a real web index and ask judges to rate which image is more attractive in the pair by choosing one of five labels - ``left better", ``left slightly better", ``equal", ``right slightly better", and ``right better". The ``slightly better" is added according to the feedback from judges who feel this additional label provides comfortable flexibility. We call our side-by-side web data (SBS). The judge only needs to make a relative comparison between two images without the necessity to assign a numerical score to an image. The judgment policy is purely based on human perception: ``select which image in the pair looks more attractive based on your perception". This task can be easily completed by the judges without much image domain knowledge, and is more intuitive. We propose an efficient image pair sampling methodology to generate image pairs. Then, we build a novel deep attractiveness rank network (DARN) which is able to simultaneously learn the score mean and variance of each image,and the underlying criteria the judges use to label each pair, from these side-by-side rated image pairs. The DARN is inspired from rank net~\cite{Datta06}, but has very different network structure. The DARN model can be trained on side-by-side data with arbitrary label dimension. The DARN also has the flexibility to be trained on the data set with absolute rating via appropriate pre-processing. Based on DARN, we also propose DARN-V2 model which adapts to individual judge's personal preference. Later, We demonstrate that the attractiveness of search results can be significantly improved by properly utilizing the attractivness information in a real commercial search engine. We evaluate our
model against a few state-of-art models on our side-by-side web test data and AVA data set and demonstrate that our model outperforms on side-by-side labeled data and is comparable on data with absolute labels without fine tuning. The models trained on the AVA data do not perform well on the images collected from real web image index, and thus may cause potential problem if applying those pre-trained models in image retrieval application directly. We plan to release our data to make it available for training and evaluating image quality models for image retrieval application.

\section{Data Collection}
\begin{figure}[t]
\begin{center}
\includegraphics[width=0.8\columnwidth]{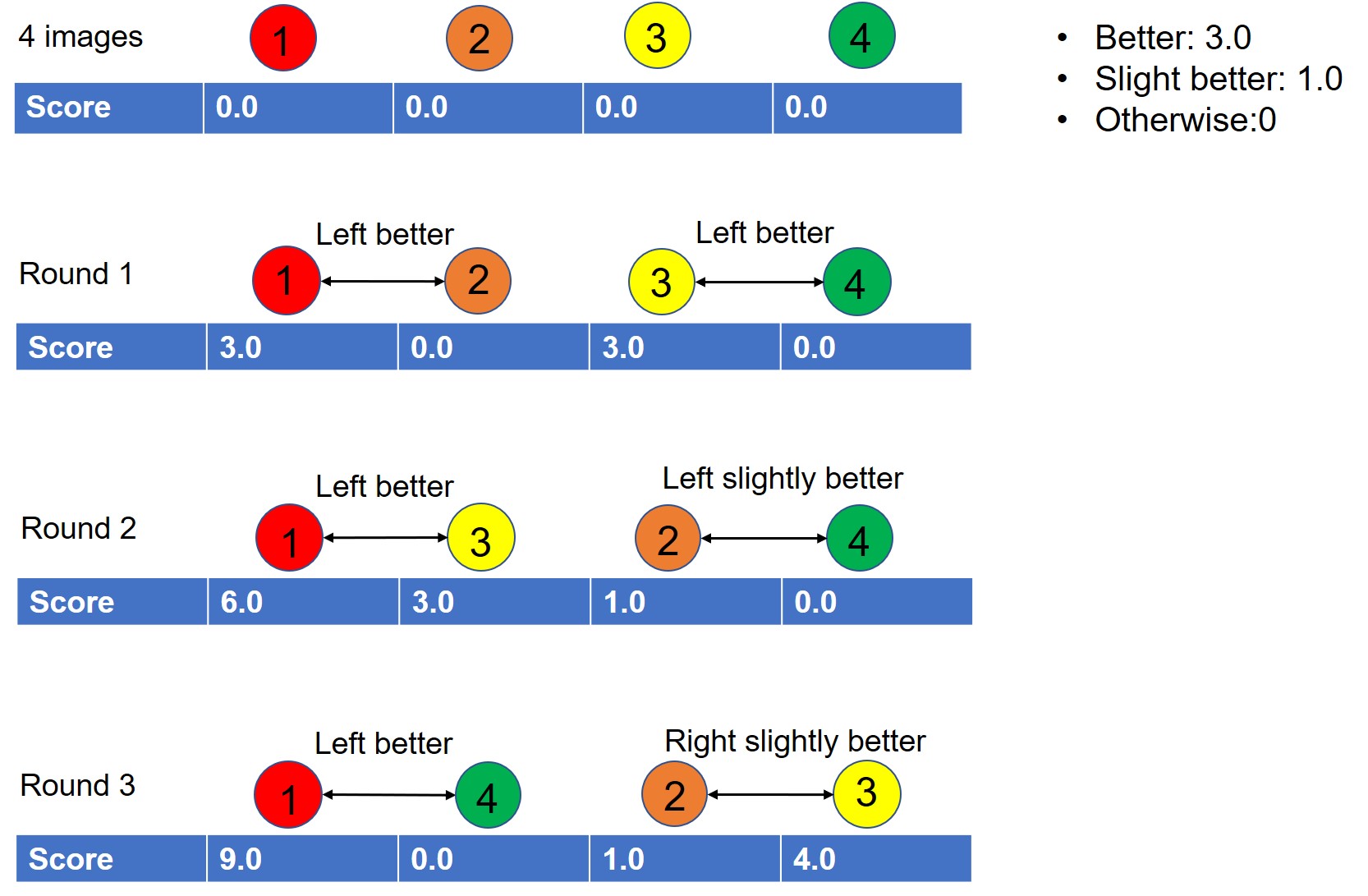}
\end{center}
   \caption{Graphical illustration of swiss tournament sampling method. Circles with different colors denotes different images}
\label{fig:swisstournament}
\end{figure}

\begin{figure}[t]
\begin{center}
\includegraphics[width=0.8\columnwidth]{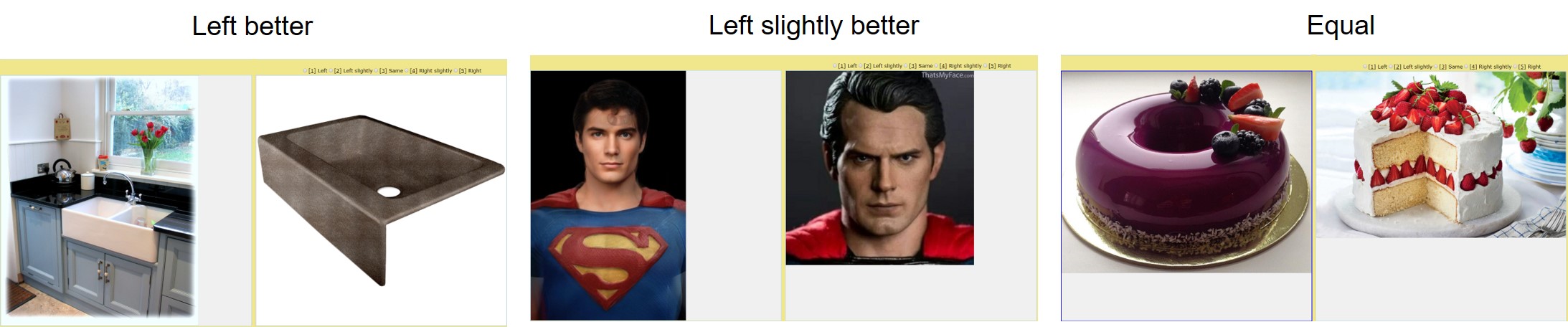}
\end{center}
   \caption{Example of our side-by-side image attractiveness labeling. From left to right are three image pairs where most judges label as ``left better", ``left slightly better" and ``equal".}
\label{fig:labelsample}
\end{figure}

\begin{figure*}
\begin{center}
\includegraphics[width=2\columnwidth]{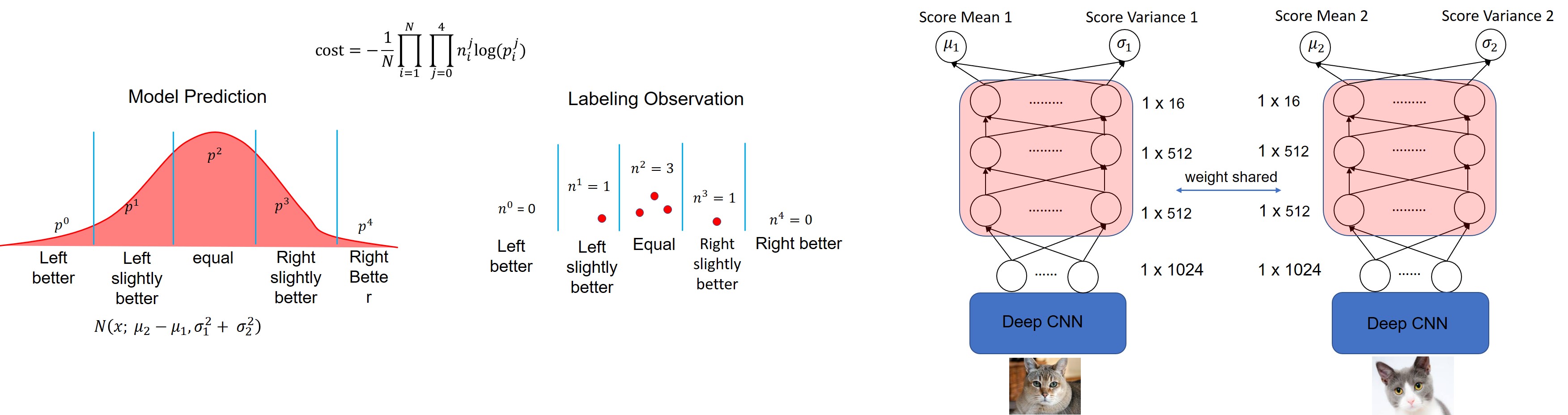}
\end{center}
   \caption{The DARN learns a mapping from an image to its score mean and variance, and a mapping from score difference to the human label. \textbf{Left column}: The distribution of the score difference between images in one pair and the corresponding human pairwise labeling observation. The boundaries are used to map each pair to a label according to their score difference.\textbf{Right column}: The deep attractiveness rank net. The high level features of each image is extracted by a deep CNN. A pairwise DNN is used to map these features to a score mean and variance. The decision boundaries are learned simultaneously.}
\label{fig:basemodel}
\end{figure*}

We collect a large data set of image pairs each labeled by human judges. We first select 10k queries. For each query, we scrape top 20 images and 10 random images from next 1000 images returned from Bing. For each query, we have at most 30 images. 

One challenge of collecting side-by-side pairs is that $N$ images will generate $ \mathcal{O}(N^2)$ pairs if we exam all possible pairs. Here, we use a pairing method resembling a swiss tournament system. During comparison, the image will get a score of 3.0 if it gets a ``better" label, and 1 point for ``slightly better" better, 0 otherwise. After each round, we will regroup images according to their current scores. We first pair images in the group with highest score, and then the ones with second highest score, etc. The pairing is only performed between images belonging to the same query, and required not to generate duplicate pairs. Figure~\ref{fig:swisstournament} shows the process of running three rounds of the swiss tournament. For more detail, please see \cite{swisslink}.  Each pair will be shown to judges in normal and flipped orientation to account for left-right bias.

We select 500 queries each as our validation and test queries. The remaining 9K queries serve as training queries. We run five tournament rounds on the validation and test queries to obtain our validation and test set of image pairs, and two rounds on the training queries to obtain the training pairs. Each round generates $N/2$ image pairs from $N$ images. Each judge will compare only two images side-by-side and rate relative attractiveness using one of five labels - ``left better", ``left slightly better", ``equal", ``right slightly better", and ``right better". The judge is shown only the image pairs without seeing the corresponding query. Figure~\ref{fig:labelsample} shows a sample of image pairs to be rated by the judges. 

In practice, we found that the model trained on image pairs obtained from only two rounds of rating, that is only $\mathcal{O}(N)$ pairs, already gave a good attractiveness ranking result.

\begin{figure}[t]
\begin{center}
\includegraphics[width=0.8\columnwidth]{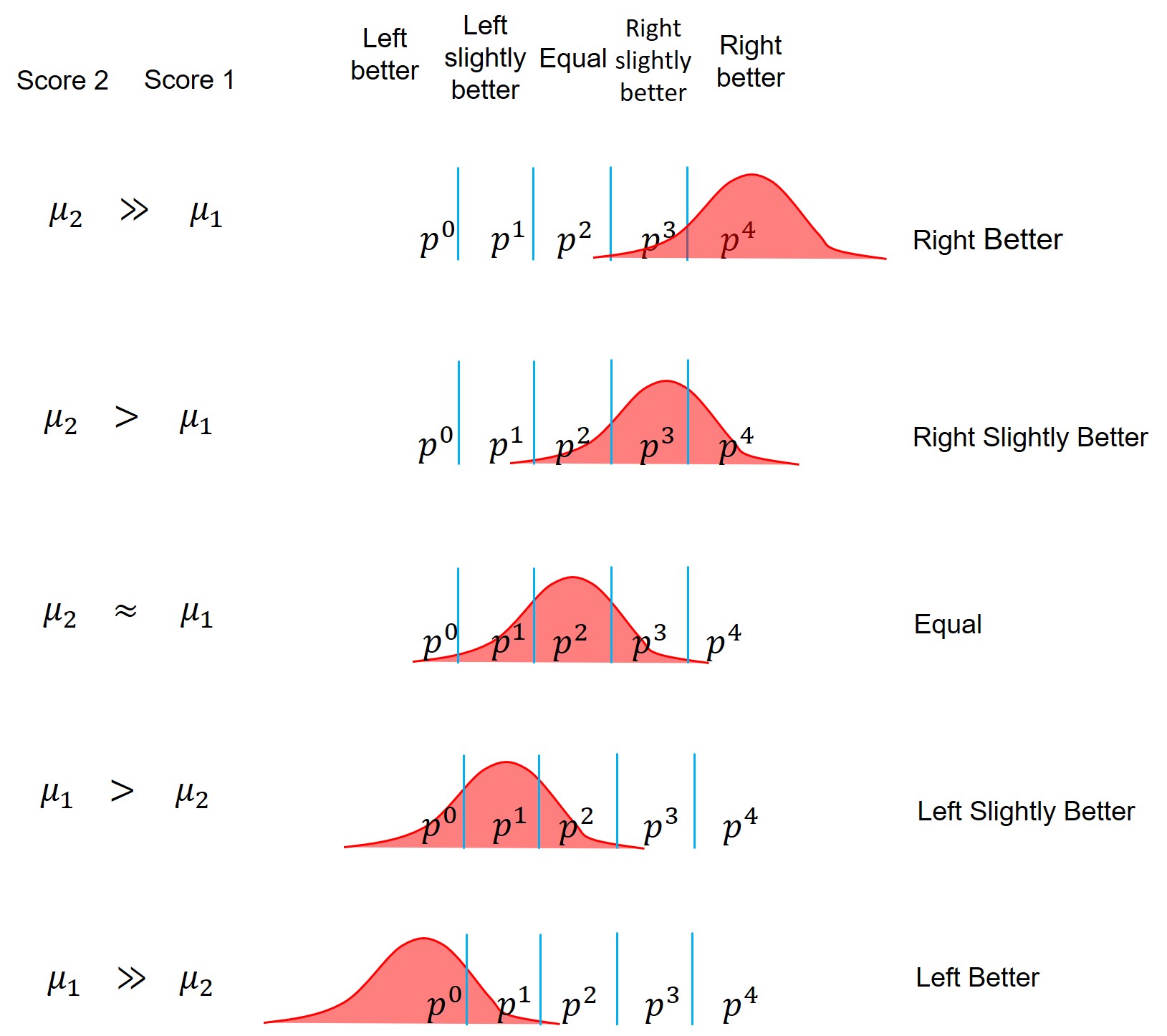}
\end{center}
   \caption{Figure shows how a label will systematically change with the score difference of an image pair. For example, in the first row, when the mean score of the right image is much higher than the left one, the distribution of score difference will shift to the right and result in higher probability for label 'right better', i.e., largest $p^4$ value.}
\label{fig:labelmapping}
\end{figure}

\section{Model Implementation}
\subsection{Bayesian Ranking Cost Function}
We would like to train a machine learning model with pairs of images $[x_1,x_2]$, together with human preference for each pair $Y$. The goal is to learn a model $f:\mathbb{R}^d \mapsto \mathbb{R}$ such that the images can be ranked by the attractiveness score specified by the model (i.e., $f(x_1) > f(x_2)$ indicates image $x_1$ is more attractive than $x_2$). Figure ~\ref{fig:basemodel} graphically illustrates the deep attractiveness rank net framework. The model is designed to predict the mean and variance of the attractiveness score for an image, and the decision boundary that specifies how differences in these distributions correspond with judge preferences.

Let us define $\mu = \mathrm{E}[f(x)]$ and $\sigma^2 = \mathrm{Var}[f(x)]$ as the mean attractiveness score and variance for an image. For an image pair $[\mathbf{x_i}:(\mu_i, \sigma_i), \ \mathbf{x_j}:(\mu_j, \sigma_j)]$, define $\mathrm{P_{ij}}(\mathbf{x_i}\leftrightarrow\mathbf{x_j} = y)$ as the posterior probability that the image pair is labeled as $y$. In our five-level labeled data, we have 
\begin{IEEEeqnarray}{rCl}
\sum\limits_{y=0}^{4}\mathrm{P_{ij}}(\mathbf{x_i}\leftrightarrow\mathbf{x_j} = y) = 1
\end{IEEEeqnarray}
Intuitively, the attractiveness score should have at least the following properties:
\begin{itemize}
\item The more attractive the image, the higher the mean score.
\item Labeling should be consistent with score difference. For example, a pair should be more likely to be labeled with 'left better' when $\mu_i \gg \mu_j$, and 'equal' when $\mu_i \approx \mu_j$.
\end{itemize}
Assume each image can be rated by a large number of experts who have extremely high confidence of attractiveness rating.  According to central limit theorem, the attractiveness scores received by each image will follows a normal distribution $\mathcal{N}(x;\mu,\sigma)$. For the two images $x_i$ and $x_j$, the score difference is also a normal distribution
\begin{IEEEeqnarray}{rCl}
 \mathcal{N}(x;\mu_i - \mu_j,\sigma_i^2 + \sigma_j^2)
\end{IEEEeqnarray}
as shown in left panel of Figure~\ref{fig:basemodel}.
The model learns four boundaries which are used to map each pair to a label according to their score difference. We define the four boundaries as $\{b_i\}_{i = 0}^3$. Let $p_i^j$ denote the probability that the $i_{th}$ pair is labeled as $j$ (indexing \{left better, left slightly, equal, right sightly, right better\} as \{0, 1, 2, 3, 4\}), and $\bigtriangleup\mu_i$ and $\bigtriangleup\sigma_i$ as the mean and variance of the score difference of $i_{th}$ pair. So, we have

\begin{IEEEeqnarray}{rCl}
p_i^j = \int\limits_{b_{j - 1}}^{b_{j}}\mathcal{N}(x;\bigtriangleup\mu_i, \bigtriangleup\sigma_i)dx
\end{IEEEeqnarray}

$p_i^j$ is the probability of the $i_{th}$ pair labeled as $j$ and is represented by the area under the normal distribution of the score difference of the $i_{th}$ pair between boundary $b_{j - 1}$ and $b_j$. Figure~\ref{fig:labelmapping} illustrates how the label will systematically change with the score difference. For example, when the mean score of right image is much higher than the left one, the distribution of score difference will shift to right and result in higher probability for label 'right better'. When the score difference between right and left becomes smaller, the distribution shifts to the left and causes the labeling to change. As a result, when the scores of two images are equal, the area in the middle bucket is largest, meaning that the pair will be most likely be labeled as 'equal'.

Let $n_i^j$ indicate the number of judges labeling $i_{th}$ pair as label $j$. Thus, we define a log maximum likelihood cost function as 

\begin{IEEEeqnarray}{rCl}
cost & = -& \log \prod\limits_{i = 1}^{N}\prod\limits_{j = 0}^{4}(p_i^j)^{n_i^j}\\
& = -& \sum\limits_{i = 1}^{N}\sum\limits_{j = 0}^{4}n_i^j\log (p_i^j)
\end{IEEEeqnarray}
where N is the number of pairs.

\subsection{Deep Attractiveness Rank Network (DARN)}
The right panel of Figure~\ref{fig:basemodel} shows the structure of the deep attractiveness rank net. The image features are extracted from the final 1024-dimensional activations of a deep CNN which is very similar to the Inception-BN model \cite{Ioffe2015} pretrained on Imagenet data \cite{Deng2009}. The DNN features are then fed to three fully connected layers to generate a mean score and variance for each image. For each hidden layer, we have 
\begin{IEEEeqnarray}{rCl}
\mathbf{h}^{(n)} & = & g^{(n)}\Big(\mathbf{W}^{(n)\top}\mathbf{h}^{(n - 1)} + \mathbf{b}^{(n)}\Big)
\end{IEEEeqnarray}
where $\mathbf{h}^{\{n\}}$ denotes the activation value of each node in the $n_{th}$ layer, $\mathbf{W}^{\{n\}}$ the weights of the $n_{th}$ layer, and $\mathbf{b}^{\{n\}}$ the bias . We apply ReLu activation function $g(x)$ on each layer before feeding to the next layer. The network will output a mean score node
\begin{IEEEeqnarray}{rCl}
\mathbf{\mu}& = & g\Big(\mathbf{W}_{\mu}^{T}\mathbf{h}^{(N)} + \mathbf{b}_{\mu}\Big)
\end{IEEEeqnarray}
and a standard deviation node
\begin{IEEEeqnarray}{rCl}
\mathbf{\sigma}& = & g\Big(\mathbf{W}_{\sigma}^{T}\mathbf{h}^{(N)} + \mathbf{b}_{\sigma}\Big)
\end{IEEEeqnarray}
We use dropout to avoid overfitting with a dropout rate 0.5 and use ReLU as the activation function. 

By minimizing the cost function, the deep neural network learns a mapping from each DNN feature to a score mean and variance, and a mapping from score difference between two images to the labels by jointly learning the neural network weights $\{\mathbf{W}, \mathbf{\mu}, \mathbf{\sigma}\}$ and the four labeling boundaries $\{b_i\}_{i = 0}^3$. The neural network is trained on pairs and shares the weights for each image of a pair. It only updates parameters via gradient descent backpropagation after a batch of pairs of images are fed to the network. We let the model learn the score mean and variance of each image and the underlying criteria the judges use to label each pairs. We let humans perform an easier and less subjective side-by-side comparing task and leave the hard learning task to a machine learning model. 

This model can be easily extended to side-by-side data with arbitrary label dimension by reducing or increasing the number of boundaries. For example, a binary version DARN can be trained on binary side-by-side data using one decision boundary.
\subsection{DARN-V2}
In DARN, we assume all judges share the same decision criteria. As a result, the decision boundary $\{b_i\}_{i = 0}^3$ is shared across all judges. However, each judge has personal bias, and thus resulting in a slightly different decision boundary. We capture individual judge preference by defining the $i_{th}$ decision boundary for judge $r$ as
\begin{IEEEeqnarray}{rCl}
b_i^r = b_i^0 * \gamma^r
\end{IEEEeqnarray}
where$\{b_i^0\}_{i = 0}^3$ is the base decision boundary shared across all judges and $\gamma^r$ is a personal preference scale factor for judge $r$. Since each judge rates one pair at most once, the loss function becomes 
\begin{IEEEeqnarray}{rCl}
cost & = -& \log \prod\limits_{i = 1}^{N}\prod\limits_{j = 0}^{4}\prod\limits_{r \in \mathrm{R_i}}(p_i^{(j,r)})\\
& = -& \sum\limits_{i = 1}^{N}\sum\limits_{j = 0}^{4}\sum\limits_{r \in \mathrm{R_i}}\log (p_i^{(j,r)})
\end{IEEEeqnarray}
where $\mathrm{R_i}$ denotes all judges rating $i_{th}$ pair, and $p_i^{(j,r)}$ the probability the $i_{th}$ image pair is labeled as $j$ by judge $r$

\begin{IEEEeqnarray}{rCl}
p_i^{(j,r)} & = & \int\limits_{b_{j - 1}^r}^{b_{j}^r}\mathcal{N}(x;\bigtriangleup\mu_i, \bigtriangleup\sigma_i)dx
\end{IEEEeqnarray}
The model jointly learn the base decision boundary and personal scale factor for each judge.

\section{Evaluation}
In this section, we evaluate our model performance on side-by-side web data (SBS) and the AVA data set. We compare our model with state-of-art models. In this paper, we mainly focus on the comparison against Kong \etal~\cite{Kong2016}, and Talebi and Milanfar~\cite{Talebi2017}. Since our side-by-side data and model are fundamentally different from others, a full fair comparison is a bit difficult. For example, it is not feasible to train other competitive models on the side-by-side data, and training DARN on the AVA data requires appropriate pre-processing. The details of the evaluation is demonstrated in the following section.

\subsection{On SBS Web Data}
\begin{table}
\begin{center}
\begin{tabular}{|l|c|c|}
\hline
Data &\# of Queries & \# of Image Pairs\\
\hline\hline
Training Data & 9K & 298K\\
Validation Data & 0.5K & 43K\\
Test Data & 0.5K & 43K\\
\hline
\end{tabular}
\end{center}
\caption{Data size used for training and testing the model}
\label{table:SBSData}
\end{table}

\begin{table}
\begin{center}
\begin{tabular}{|l|c|c|c|}
\hline
Model & Train&Test& SBS-Acc\\
\hline\hline
Majority Vote& - & SBS & 39.08\%\\
DARN & SBS & SBS & 67.78\%\\
\textbf{DARN-V2} & \textbf{SBS} & \textbf{SBS} & \textbf{68.95}\% \\
\hline
\end{tabular}
\end{center}
\caption{Our model's \textbf{five-way} prediction accuracy on side-by-side web (SBS) test data}
\label{table:SBSfiveway}
\end{table}

\begin{table}
\begin{center}
\begin{tabular}{|l|c|c|c|}
\hline
Model & Train&Test& SBS-Acc\\
\hline\hline
Reg+Rank+Att+Cont~\cite{Kong2016} & AVA & SBS & 57.86\%\\
NIMA(Inception-V2)~\cite{Talebi2017} & AVA & SBS & 57.92\% \\
\textbf{DARN-Binary} & \textbf{AVA} & \textbf{SBS} & \textbf{58.78}\%\\
\hline\hline
\textbf{DARN-V2} & \textbf{SBS} & \textbf{SBS} & \textbf{74.20}\%\\
\hline
\end{tabular}
\end{center}
\caption{\textbf{Binary} prediction performance comparison on SBS web test data using the AVA training data.}
\label{table:SBSbinary}
\end{table}

\noindent \textbf{Side-by-Side Web Data.} Table~\ref{table:SBSData} gives a summary of our side-by-side web data (SBS) that we use to train and test our model. We use the images associated with 9k queries to generate pairs for training, 500 queries for validating and 500 queries for testing. We run the swiss tournament sampling method five rounds on the test query set to get test pairs, and two rounds on training query to obtain training pairs. We only use the pairs in which more than half of the judges agree on one of the five labels. Finally, we have about 43K images pairs as test data and 300K pairs as training data.  Each image pair is judged by about five judges, resulting in about 1.5M judged pairs.\\

\noindent \textbf{Five-Way Prediction Accuracy on Side-by-Side Wed Test Data.} The true label of each pair is taken as the majority vote of all judges' rating on this pair $(y = \argmax\limits_j n^j)$. In DARN, the predicted label for this pair is the label which is most likely to be assigned to the pair $(\hat{y} = \argmax\limits_j p^j)$. We train DARN on training pairs using DNN features obtained from a pre-trained deep CNN network similar to the structure of ~\cite{Ioffe2015}. Size information, $(\mathrm{\frac{width}{max\  length}}$, $\mathrm{\frac{height}{max\ length}})$, is also appended to the DNN feature before feeding to the next level network.

In the DARN-v2 model, each judge will have personal decision boundaries. The prediction for judge $r$ on a pair is denoted as $\hat{y^r} = \argmax\limits_j p^{(j,r)}$. The final prediction on this pair is the majority vote of all judges' predictions on this pairs. Table~\ref{table:SBSfiveway} shows the five-way prediction accuracy on the test data set. The five-way prediction accuracy of DARN and DARN-v2 are 67.78\% and 68.95\%, respectively. 

We use a majority vote predictor as a baseline. The majority vote rates all pairs as the most popular human label across all pairs. The DARN model is significantly better than the majority vote predictor.

\noindent \textbf{Binary Prediction Accuracy on Side-by-Side Web Test Data.} Most existing image quality classifiers do not provide five-way side-by-side prediction. In order to compare the performance against state-of-art models on this SBS data set, we group the human labels into two categories. The label is converted to 1 for ``right better" and ``right slightly better", and to 0 otherwise. The DARN-V2 achieves 74.20\% accuracy, if we simply convert five-way prediction to binary prediction as shown in the last row of the Table~\ref{table:SBSbinary}. 

Each model can generate an attractiveness score. If the score of left image is not smaller than the right image, the prediction is 0, otherwise 1. So, we can compare different models using binary side-by-side accuracy. This metric is also invariant to monotonic transformations of the attractiveness score predictions.

We trained a binary DARN model (DARN-Binary) on the AVA data set. The DARN-binary is constructed simply by replacing the four boundaries with one boundary. Since AVA data only provides absolute score for each image, we generate pairs by the following simple pre-processing procedure. We randomly sampled about 20K images from the original training data. For each image, we randomly selected 50 other images to get pairs without duplicates. If the left image score is not smaller than the right one, we label this pair as 0, otherwise 1. So, we got about 1M synthesized image pairs. We trained our binary DARN model on these synthesized side-by-side labeled data and test it on our SBS web test data.

Table~\ref{table:SBSbinary} shows the binary prediction performance on SBS data of different models. We mainly compare with Reg+Rank+Att+Cont~\cite{Kong2016}, and NIMA(Inception)~\cite{Talebi2017} models which are trained on the AVA data set.  We show that the models trained on the AVA do not have very good performance on the SBS web test data, and the DARN outperforms the others. This is probably because our data is from a real image search index and much more diverse than the AVA. The two data sets have quite a different distribution.
\begin{figure*}
\begin{center}
\includegraphics[width=1.4\columnwidth]{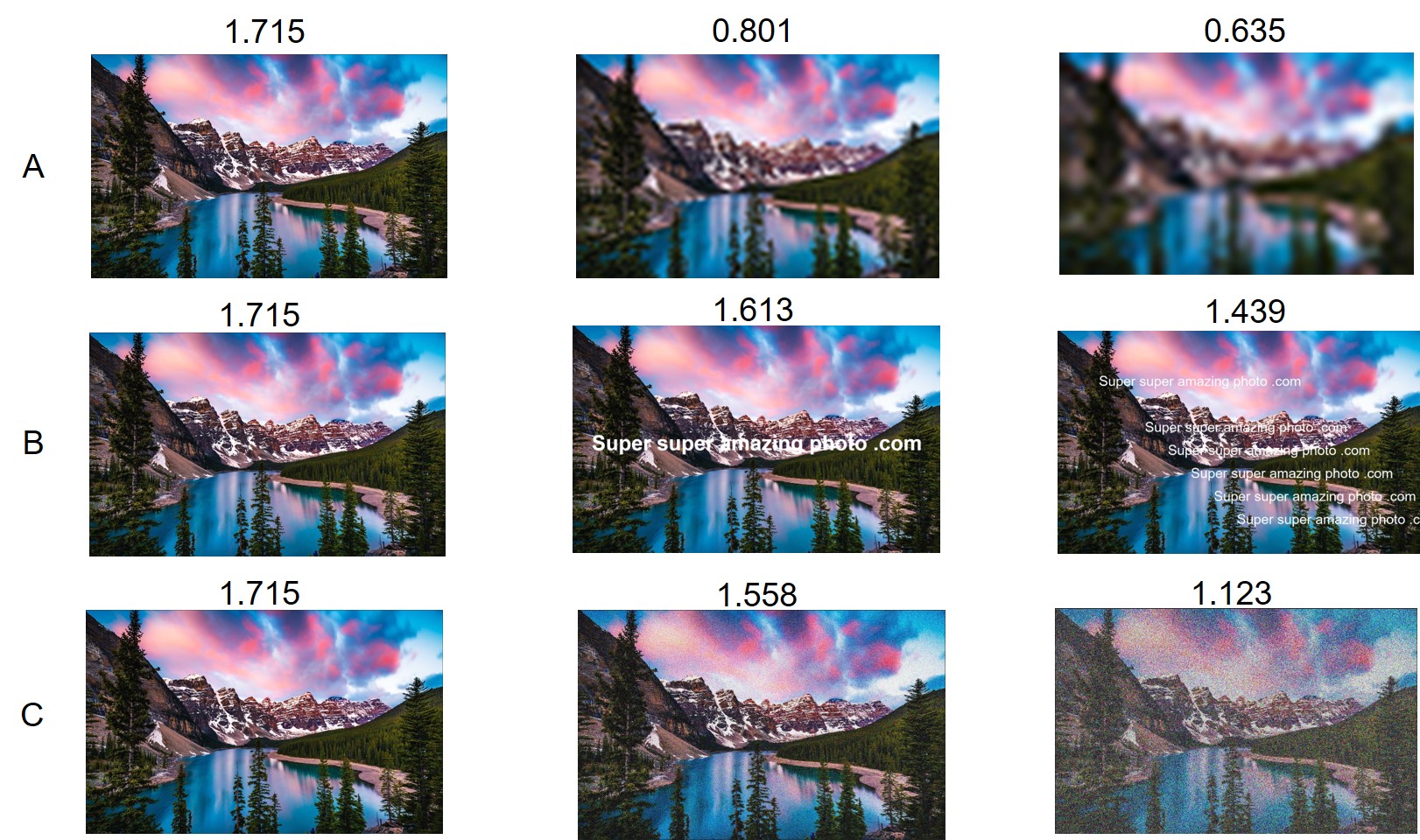}
\end{center}
   \caption{Image score decreases with image quality degeneration. Image score decreases as the resolution is reduced \textbf{(A)}, more watermarks are applied \textbf{(B)}, and more Gaussian noise is added \textbf{(C)}.}
\label{fig:scoreinsight}
\end{figure*}

\begin{figure*}
\begin{center}
\includegraphics[width=1.4\columnwidth]{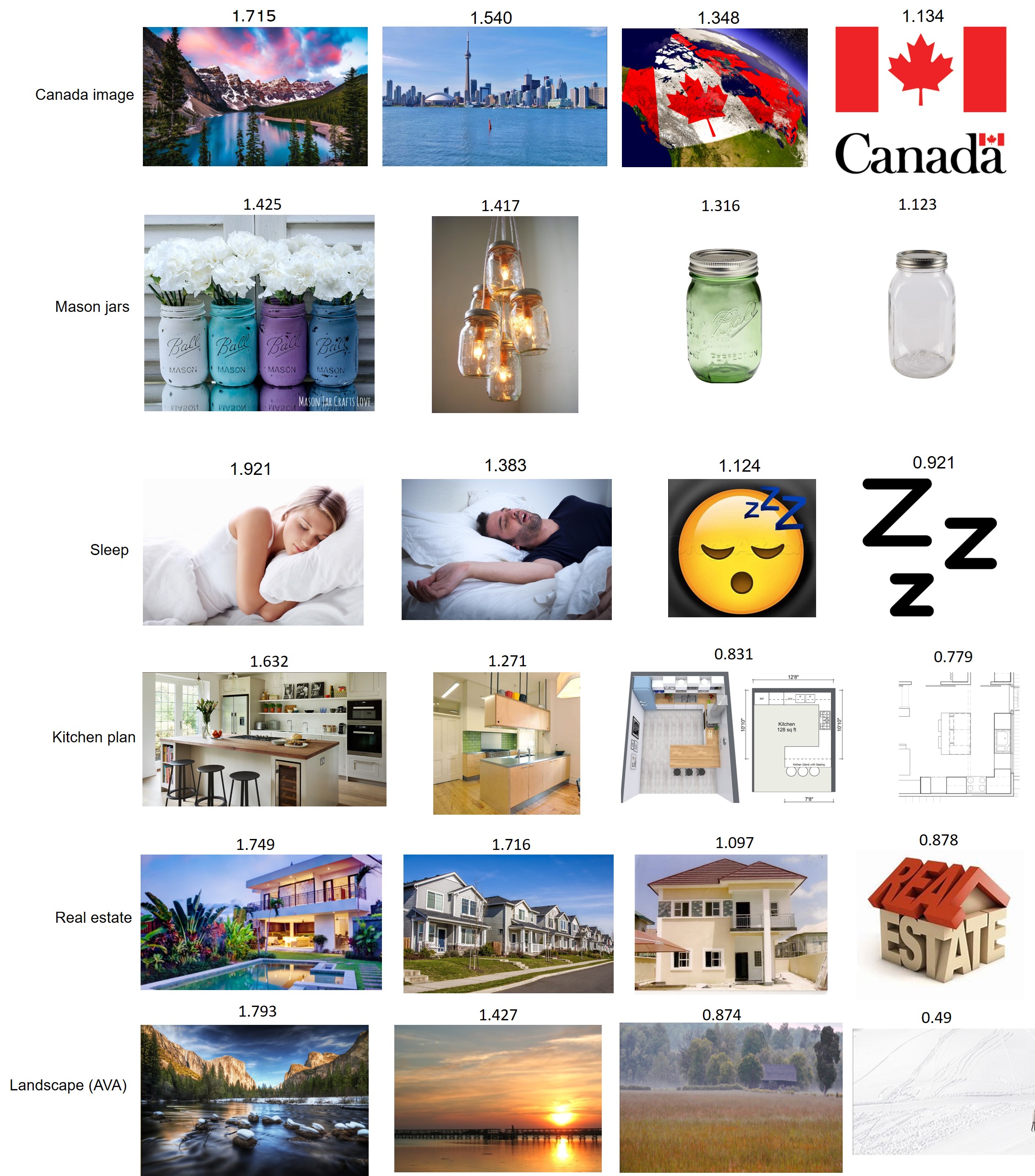}
\end{center}
   \caption{Rank images by attractiveness score for different queries. Each row includes four images associated with the corresponding query/categories. The number above each image is the attractiveness score. Only the last row is from AVA~\cite{Murray2012} data set. All other examples are scraped from web image index.}
\label{fig:rankimagebyscore}
\end{figure*}

\begin{figure*}
\begin{center}
\includegraphics[width=1.7\columnwidth]{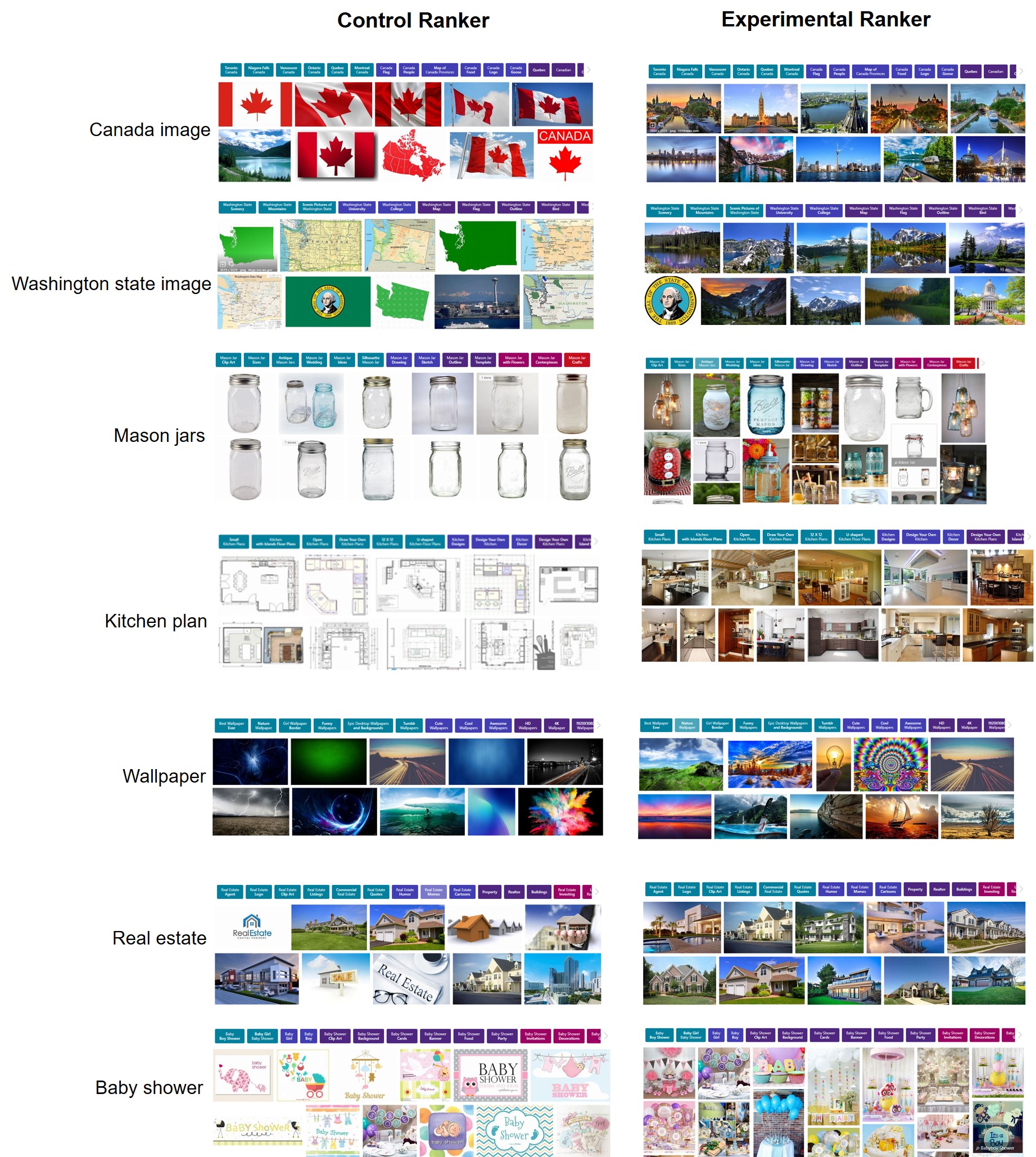}
\end{center}
   \caption{Search results boost image attractiveness significantly without deteriorating image relevance. The left and right columns represent images from old image search ranker and new ranker, respectively. Each row is the search results of two search rankers for a specific query. Images are crops of top two rows from the results returned by search ranker.}
\label{fig:rankercompare}
\end{figure*}

\begin{table}
\begin{center}
\begin{tabular}{|l|c|c|c|}
\hline
Model & Train & Test & SRCC \\
\hline\hline
AlexNet\_FT\_Conf~\cite{Kong2016} & AVA & AVA & 0.481\\
Reg~\cite{Kong2016} & AVA & AVA & 0.500 \\
Reg+Rank~\cite{Kong2016} & AVA & AVA & 0.513\\
Reg+Rank+Att+Cont~\cite{Kong2016} &AVA & AVA &  0.558\\
NIMA(MobileNet)~\cite{Talebi2017} &AVA & AVA &  0.510\\
\textbf{NIMA(Inception-V2)}~\cite{Talebi2017} &AVA & AVA & \textbf{0.612}\\
DARN-Binary & AVA & AVA & 0.516\\
\hline
\end{tabular}
\end{center}
\caption{Performance comparison on AVA~\cite{Murray2012} test data using AVA~\cite{Murray2012} training data}
\label{table:AVAAVA}
\end{table}

\subsection{On AVA Data Set}
\begin{table}
\begin{center}
\begin{tabular}{|l|c|c|c|c|}
\hline
Model & Train & Test & SRCC\\
\hline\hline
Reg+Rank+Att+Cont~\cite{Kong2016} & AADB & AVA & 0.157\\
NIMA(Inception-V2)~\cite{Talebi2017} & TID2013 & AVA & 0.087 \\
NIMA(Inception-V2)~\cite{Talebi2017} & Live & AVA & 0.200\\
\textbf{DARN-V2} & \textbf{SBS} & \textbf{AVA} & \textbf{0.245}\\
\hline
\end{tabular}
\end{center}
\caption{Performance comparison on AVA~\cite{Murray2012} test data using other training data, \eg AADB~\cite{Kong2016}, TID2013\cite{Ponomarenko2013}, LIVE\cite{Ghadiyaram2016}, and our side-by-side web data (SBS).}
\label{table:otherAVA}
\end{table}

\noindent\textbf{AVA Data Set.}
The AVA data contains about 255K images of which 20K are used as a test set. Each image is rated by about 200 people, resulting in about 50M judgments, the number of which is much larger than the ones used in the side-by-side data. The rating ranges from 1 to 10. The average mean score is used as the final label for each image.

\noindent\textbf{Spearman Ranking Correlation Coefficient on AVA test data.}
We use Spearman's Ranking Correlation Coefficient (SRCC) as the metric for comparing different models on the AVA test data, because it is invariant to monotonic transformations of the attractiveness score predictions. This is particularly useful for our DARN model trained on the AVA data as we do not predict absolute human score. Table~\ref{table:AVAAVA} shows the performance of each model on the AVA test data using the AVA training data. The DARN-Binary model trained on the 1M synthesized pairs, without using fine tuning, achieves comparable performance against a few state-of-the-art models, even though it is trained using synthesized side-by-side data without using any absolute human score. Table~\ref{table:otherAVA} shows how the models perform on the AVA test data when trained on a different data set. It shows that DARN trained on SBS web data has better transferability.

\section{Application}
In this section, we first demonstrate some basic factors that affect the image attractiveness. Then, we show a few examples of ranking real web images by the attractiveness score. Finally, we show that the attractiveness of multimedia search results can be significantly improved by utilizing the attractiveness information.
\subsection{Rank Images by Attractiveness Score}
Figure~\ref{fig:scoreinsight} shows how the image score decreases as basic image quality degenerates. The results show that the image attractiveness decreases as the image becomes more blurred, more noisy, and has more watermarks. This observation is consistent with human visual perception.

Figure~\ref{fig:rankimagebyscore} shows the ranking of images according to their attractiveness score.  Each row corresponds to the ranking of images associated with one query. We can see that our model can rank image attractiveness for a variety of query images remarkably well. The model score captures complicated image aesthetic content, and prefers images having rich content, colorful appearance, high resolution, and better aesthetic property.

\subsection{Effects on a Commercial Search Engine}

We also incorporate the attractiveness score information into a commercial image ranker used in real commercial search engine Bing. The baseline search ranker is trained against a hybrid metric which emphasizes factors including but not limited to image relevance, image quality and page quality. However, the baseline image quality metric does not have differential information about image attractiveness. This leads to a lot of queries returning relevant but not appealing images as shown in the left column in Figure ~\ref{fig:rankercompare}. By utilizing the image attractiveness information obtained from our model, we designed a new hybrid metric. We used this new metric to train a new image search ranker which optimized both relevance and attractiveness simultaneously. The right column in the Figure~\ref{fig:rankercompare} shows that the new ranker boosts the image attractiveness significantly. It also wins in side-by-side competition. The details of how we design the new metric, train the image search ranker and evaluate the online results are beyong the scope of this paper. We will write those details in a separate paper.

\section{Summary}
\label{discussion}
In this paper, we propose a novel pairwise deep rank net trained on side-by-side five-way labeled image pairs. The raters only need to provide relative ranking between two images using a simple rating policy without the need to either assign an absolute score, or rate predefined image attributes. The task is more intuitive and less expensive compared to absolute score rating tasks. Our novel model learns the mean score and variance, and each judge's personal decision criteria simultaneously. The model has flexibility to be trained on data with absolute rating as well. We compare our model with state-of-the-art models and demonstrate competitive performance with much less judgments. We observe significant image quality ranking performance improvement when using this information in a real commercial search engine.

Our side-by-side rated data set is fundamentally different from the AVA data set. Our data set, once released, can be used to train and evaluate real web image quality models. Our model's performance can serve as a benchmark against which further image quality models can be evaluated.  

\section{Acknowledgments}
We thank Anil Akurathi, Yokesh Kumar, Yan Wang, Xi Chen, Josh Zhao, Lin Zhu, Vladislav Mokeev, Cha Zhang for valuable discussion and help, and Arun Sacheti and Richard Qian for the support of this project.
\clearpage
{\small
\bibliographystyle{ieee}
\bibliography{egbib}
}
\end{document}